\documentclass[11pt]{article}
\PassOptionsToPackage{table}{xcolor}
\usepackage[final]{acl}
\usepackage{times}
\usepackage{latexsym}
\usepackage[T1]{fontenc}
\usepackage[utf8]{inputenc}
\usepackage{microtype}
\IfFileExists{inconsolata.sty}{\usepackage{inconsolata}}{}
\usepackage{graphicx}
\usepackage{booktabs}
\usepackage{amsmath}
\usepackage{multirow}
\usepackage{amssymb}
\usepackage{textcomp}
\usepackage{enumitem}
\usepackage{colortbl}
\definecolor{groupgray}{gray}{0.90}
\usepackage{makecell}

\makeatletter
\newif\if@appendixtoc
\@appendixtocfalse
\newcommand{\appendixTOCstart}{\global\@appendixtoctrue}

\let\old@contentsline\contentsline
\renewcommand{\contentsline}[4]{%
  \if@appendixtoc
    \old@contentsline{#1}{#2}{#3}{#4}%
  \fi
}
\makeatother

\title{Skill or Skip? Learning Selective Skill Invocation in Agentic Tasks via Dual-Granularity Preference Learning}

\author{
\textbf{Chishui Chen\textsuperscript{1,2}}\thanks{Equal contribution.}
\quad
\textbf{Jiaye Lin\textsuperscript{2}}\footnotemark[1]\thanks{Corresponding author.}
\quad
\textbf{Te Sun\textsuperscript{3}}\footnotemark[1]
\quad
\textbf{Yi Yang\textsuperscript{2,4}}\footnotemark[1]
\\
\textbf{Junxi Wang\textsuperscript{1}}
\quad
\textbf{Cong Qin\textsuperscript{2,5}}
\quad
\textbf{Yangen Hu\textsuperscript{2}}
\quad
\textbf{Lu Pan\textsuperscript{2}}
\quad
\textbf{Ke Zeng\textsuperscript{2}}
\\[2pt]
\normalfont
\textsuperscript{1}Fudan University
\quad
\textsuperscript{2}Meituan LongCat Interaction Team
\quad
\textsuperscript{3}Shanghai Jiao Tong University
\\
\textsuperscript{4}Nanjing University
\quad
\textsuperscript{5}Peking University
\\
\texttt{cschen26@m.fudan.edu.cn}
\quad
\texttt{jiaye.lin22@gmail.com}
}

\begin{document}
\maketitle

\begingroup
\renewcommand{\thefootnote}{}
\footnotetext{Work done during an internship at Meituan.}
\endgroup

\begin{abstract}
Agent skills are callable procedural modules that provide reusable knowledge and execution policies for complex agentic tasks.
However, existing methods mainly focus on selecting relevant skills or improving the skills themselves, while overlooking whether a relevant skill should actually be invoked at the current decision point.
Unhelpful invocations may introduce irrelevant context and disrupt an otherwise correct execution process.
To address this issue, we propose \textbf{SelSkill}, a dual-granularity preference-learning framework for selective skill invocation.
SelSkill formulates skill use as a \textit{skill-or-skip} decision, uses predictive uncertainty to prioritize candidate decision points, and constructs controlled invoke-skip preference pairs from shared trajectory prefixes.
It further combines episode-level outcome preferences with step-level invocation preferences to capture both overall trajectory quality and the local effectiveness of skill invocation.
On ALFWorld with Qwen3-8B, SelSkill improves task success by \textbf{10.9} points over the skill-enabled baseline and \textbf{7.8} over No-Skill, with \textbf{29.1}-point higher execution precision.
On BFCL, task success and execution precision improve by \textbf{5.7} and \textbf{29.5} points, respectively.
Zero-shot results on Tau-bench and PopQA suggest partial transfer to unseen domains and skills.
Code: \url{https://github.com/ChenChiShui/selective-skill-invocation}.


\end{abstract}

\section{Introduction}
\label{sec:intro}

As agent systems are increasingly applied to long-horizon, highly interactive tasks, relying on the model to plan and execute from scratch at each step can underuse prior experience and lead to inefficient exploration~\citep{workflowmem}.
In this context, agent skills have received growing attention in settings such as web interaction and software engineering~\citep{ASI, claudeskills, skillsbench}. As callable procedural modules, agent skills encapsulate domain knowledge, applicability conditions, and reusable execution policies, providing agents with reusable support for complex task solving~\citep{sok, skillx}.

\begin{figure*}[t!]
    \centering
    \includegraphics[
        width=\textwidth,
        trim=383bp 524bp 1552bp 65bp,
        clip
    ]{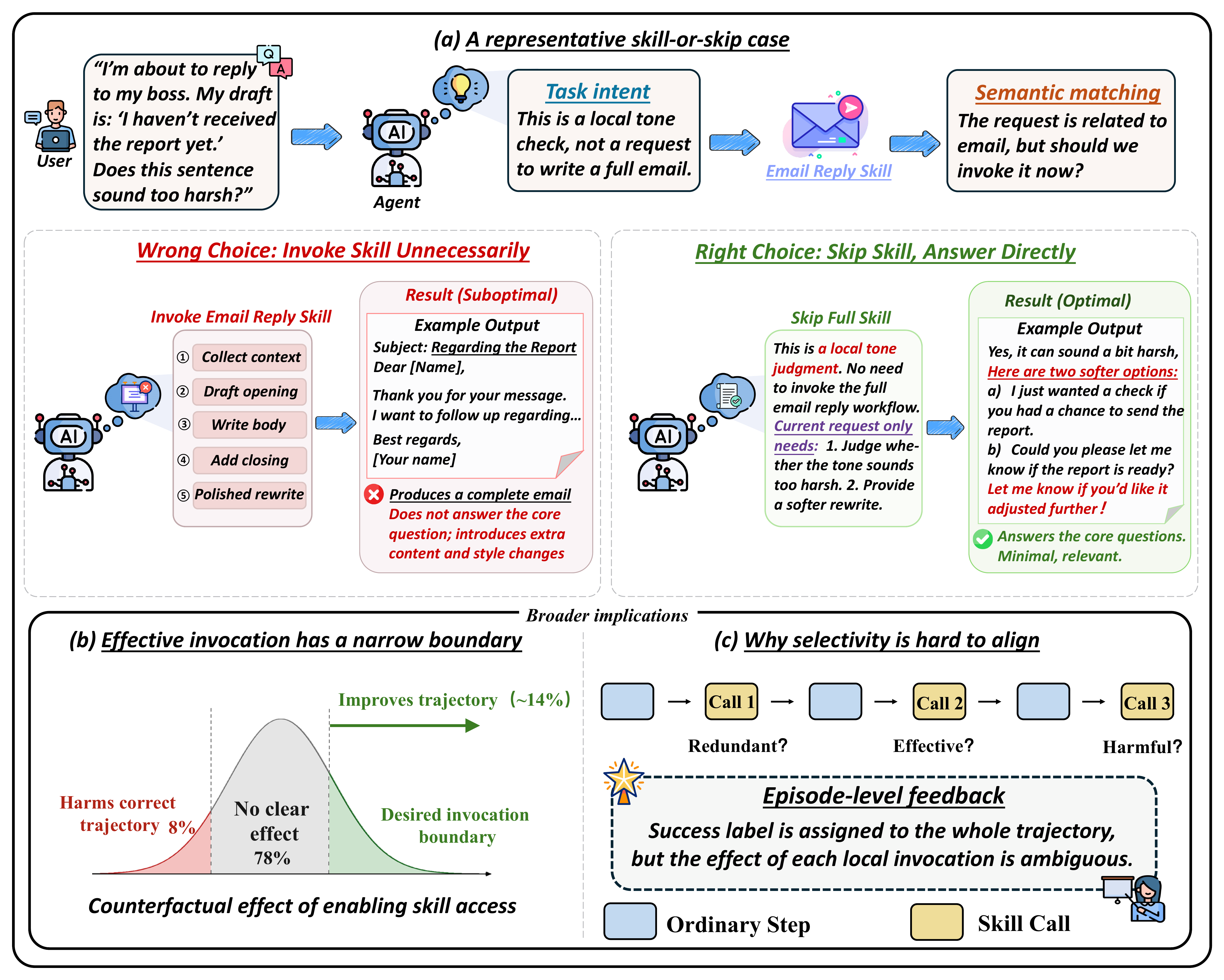}
    \caption{
    \textbf{Motivation for selective skill invocation.}
    \textbf{(a)} A representative skill-or-skip case illustrates that a relevant skill may still be unnecessary for the current request.
    \textbf{(b)} Counterfactual analysis shows that beneficial effects of skill access are concentrated in only a small fraction of paired trajectories.
    \textbf{(c)} Episode-level feedback cannot directly identify the local contribution of each invocation.
    }
    \label{fig:motivation}
    \vspace{-10pt}
\end{figure*}

Existing methods mainly focus on either identifying useful skills or improving their construction and use.
Some retrieve relevant skills from a library based on the current task context~\citep{skillrouter, skill_retrieval_augmentation}, while others construct or refine skills from external knowledge and interaction trajectories, sometimes integrating them into agent policy optimization~\citep{skillrl, d2skill}. However, these methods largely assume that relevant skills should be invoked, while overlooking a more fundamental question: even if a skill is relevant, should the agent actually invoke it at the current decision point? During task execution, unhelpful invocations may introduce irrelevant context, thereby disrupting an otherwise correct execution process. Through our analysis, we reveal two important characteristics of skills during task execution:

\textbf{(I) Highly Concentrated Skill Benefits.}
Figures~\ref{fig:motivation}(a)--(b) show that effective skill use does not mean invoking a skill whenever it appears relevant.
The case study in Figure~\ref{fig:motivation}(a) illustrates that a seemingly relevant skill call can still produce an unnecessarily broad and suboptimal response.
Across multiple benchmarks, the counterfactual results in Figure~\ref{fig:motivation}(b) show that enabling skill access improves the final outcome in only about \textbf{14\%} of paired trajectories, has no clear effect in about \textbf{78\%}, and worsens the outcome in about \textbf{8\%}.
Further analysis shows that harmful invocations are often semantically close to effective skill uses in the same context.
Thus, \textbf{\textit{the value of a skill is concentrated in a narrow set of states, requiring precise invocation rather than calling skills whenever they appear relevant.}}

\textbf{(II) Trajectory-Level Ambiguity.}
Figure~\ref{fig:motivation}(c) suggests that episode-level feedback alone makes it difficult to determine the effect of an individual skill invocation.
For example, a skill call may help complete the task, act as an unhelpful step, or have its negative effect corrected by later actions.
The final outcome does not reliably indicate whether invoking a skill was helpful at the current decision point.
This makes it insufficient to learn skill invocation only from episode-level feedback or to address this problem with simple retrieval-based rules.
Thus, \textbf{\textit{the effect of a skill invocation should be assessed at the decision-point level.}}


Therefore, learning an effective skill invocation policy requires accounting for both the concentrated benefits of skill use and the ambiguity of trajectory-level feedback.
This calls for learning signals that capture not only the overall utility of skill use for task completion, but also the local effectiveness of invoking a skill at a specific decision point.
To this end, we propose \textbf{SelSkill}, a preference-learning framework for selective skill invocation.
SelSkill uses the model's predictive uncertainty to guide the selection of candidate skill-decision points, and constructs contrastive training pairs by comparing skill invocation with skipping at these points.
Furthermore, SelSkill combines episode-level outcome preferences with step-level invocation preferences, enabling the agent to more accurately determine when to invoke a skill and when to skip it.
In summary, the main contributions of this paper are as follows:

\begin{itemize}[leftmargin=1em, itemsep=0pt, topsep=0pt]
\item \textbf{Systematic Analysis.} We provide a detailed analysis of the limitations of existing skill invocation methods and formulate selective skill invocation as a \textit{skill-or-skip} problem at each decision point, determining whether the agent should invoke a skill under the current state.    
\item \textbf{Novel Optimization Framework.} We propose \textbf{SelSkill}, a preference-learning framework for selective skill invocation, which optimizes the skill invocation policy by constructing invoke--skip contrastive pairs and combining episode-level and step-level preferences.
\item \textbf{Strong Empirical Results.} On ALFWorld, SelSkill improves task success by \textbf{10.9} points over the skill-enabled baseline and \textbf{7.8} points over No-Skill, while improving execution precision by \textbf{29.1} points. On BFCL, it improves task success and execution precision by \textbf{5.7} and \textbf{29.5} points, respectively.
\end{itemize}

\section{Related Work}
\label{sec:related}
\subsection{From Tools and Experience to Skills}
\label{subsec:tools_and_skills}
Language agents often use external tools and past experience to extend the base model.
Prior work studies API invocation, function selection, and argument generation~\citep{toolformer, gorilla}.
Later work abstracts tool chains or interaction traces into reusable procedural representations~\citep{skillcraft}.
Building on these abstractions, skills provide a compact form of reusable experience while retaining part of the executability of tools.
They package domain knowledge, applicability conditions, and executable or textual procedures~\citep{sok, claudeskills}.
They can also be organized into structured libraries to support retrieval and controlled injection during agent execution~\citep{skillx}.

\subsection{Skill Integration and Optimization}
\label{subsec:skill_integration}

Existing skill-based agent methods mainly utilize skills in three ways.
First, routing-based methods address skill selection in large libraries by matching the current context to a small set of candidate skills using routers, retrievers, or graph-based representations~\citep{skillrouter, skillnet, gos}.
Second, skill-library management methods maintain and expand the library by adding, revising, or pruning skills based on environment interaction~\citep{autoskill, trace2skill, skillpro, skillos}.
Third, building on dynamically maintained skill or experience libraries, reinforcement-learning methods use retrieved reusable knowledge to guide exploration and provide behavior priors during policy optimization~\citep{skillrl, d2skill, skill0, 2026arXiv260506130S}.
However, existing work mainly studies how to obtain, maintain, or use skills, while a relevant skill may still be unnecessary or harmful at a specific decision point, a concern also noted in recent analyses of skill-based agents~\citep{skillsbench, skill_retrieval_augmentation}.


\subsection{Selectivity in Related Agent Settings}
\label{subsec:selective_invocation}

Related studies have identified selectivity as a concern in agents that access external resources.
In tool-augmented agents, models may invoke tools when they are not helpful or fail to use tool results effectively~\citep{toollight, xu-etal-2025-alignment, when2call}.
In memory-augmented agents, retrieved experience may not match the current task context~\citep{xiong2025memory}.
Concurrent work further learns proactive retrieval over an evolving experience base through paired retrieval/no-retrieval rollouts~\citep{proactagent}.
These studies collectively suggest that external assistance should not be used indiscriminately in practice.



\section{Preliminary}
\label{sec:prelim}

We consider an agent that performs tasks in a multi-step environment.
At step $t$, the agent conditions on a trajectory prefix $h_t$ and generates an action $a_t$.
The prefix $h_t$ may include the task instruction, interaction history, environment observations, and previously returned tool or skill outputs.

In addition to ordinary actions, the agent has access to a fixed skill library $\mathcal{S}$.
Each skill $s \in \mathcal{S}$ is a callable procedural module with lightweight metadata $m_s$ and full skill content $c_s$.
The metadata includes the skill name and a short description, while the full content contains reusable knowledge, constraints, procedures, or action policies.
We denote the visible metadata of the skill library as
$M_{\mathcal{S}}=\{m_s:s\in\mathcal{S}\}$.
At decision time, the agent policy is written as
$\pi_\theta(a_t \mid h_t, M_{\mathcal{S}})$,
where $a_t$ can be either an ordinary environment action or a skill invocation.
The full skill content $c_s$ is not injected into the model context by default; it is loaded or executed only after the model explicitly invokes the corresponding skill.
Specifically, \emph{memory skills} return textual information such as strategy hints or API documentation, while \emph{executable skills} encapsulate action or tool-call sequences.
This paper does not study how to generate, modify, or improve the skills themselves.
Instead, given a fixed skill library, we study \emph{selective skill invocation}: deciding whether and when a relevant skill should intervene in a multi-step trajectory.

\begin{figure*}[h]
    \centering
    \includegraphics[
        width=\textwidth,
        trim=270bp 1045bp 485bp 235bp,
        clip
    ]{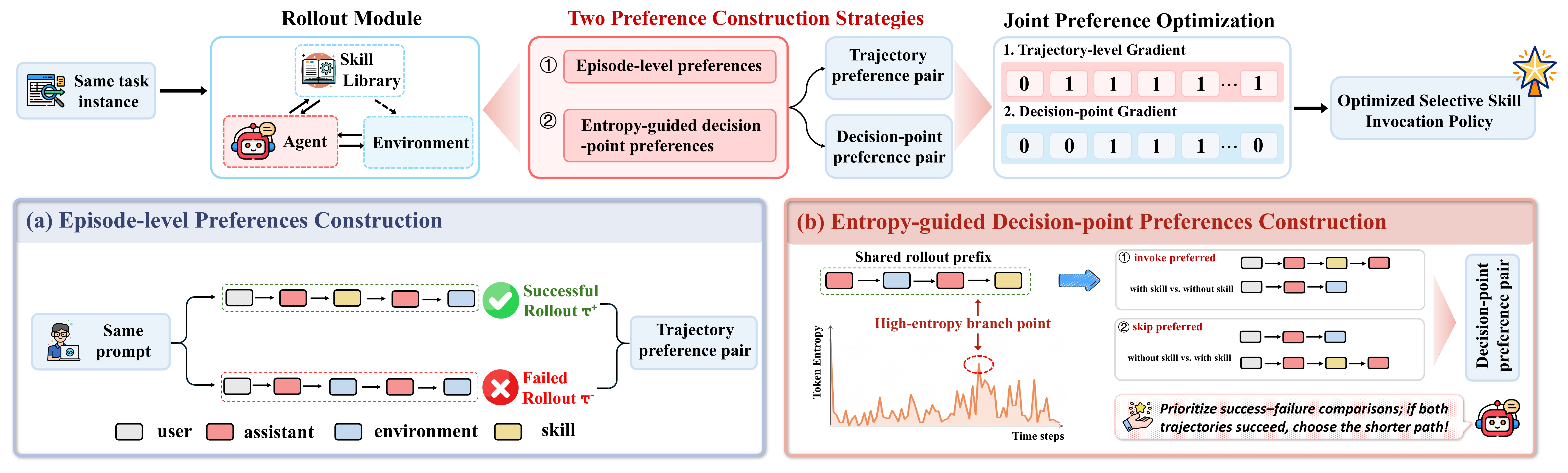}
    \caption{
    \textbf{The overview of SelSkill}.
    We construct episode-level trajectory preferences and entropy-guided decision-point preferences, 
    and jointly optimize the policy for selective skill invocation.
    }
    \label{fig:pipeline}
    \vspace{-10pt}
\end{figure*}

For each benchmark, the skill library is constructed offline from the training split and remains fixed throughout training and evaluation; construction details are provided in Appendix~\ref{app:skill_library}.

\section{Methodology}
\label{sec:method}

\subsection{Overview}
\label{subsec:method_overview}

The overview of SelSkill is illustrated in Figure~\ref{fig:pipeline}.
Our framework constructs two complementary preference signals, namely episode-level preferences and local decision-point preferences.
These two signals guide the agent to balance overall task utility with the local effectiveness of skill invocation, as detailed in the following subsections.



\subsection{Preference Construction}
\label{subsec:preference_construction}

\paragraph{Episode-level preferences.}

Episode-level preferences provide a global task-outcome signal.
For the same task, we sample multiple complete trajectories and group them according to final task success.
If one trajectory succeeds and another fails, we construct a preference pair:
\begin{equation}
(\tau^+, \tau^-),
\end{equation}
where $\tau^+$ denotes a successful trajectory and $\tau^-$ denotes a failed trajectory.
This pair indicates that the model should prefer the complete behavior sequence that solves the task.

This signal constrains the overall downstream utility of skill invocation.
It does not directly determine whether an individual skill call is necessary, but it identifies which complete trajectories are ultimately more effective.

\paragraph{Local decision-point preferences.}

A limitation of episode-level preferences is that they only provide trajectory-level success or failure feedback, making it difficult to assign credit to a specific skill invocation decision.
To directly optimize local invocation decisions, we further construct local decision-point preferences. Specifically, for each rollout, we record token-level log-probabilities during generation and compute the predictive entropy at candidate skill-decision points.
These candidate points cover uncertain states both after skill invocation and during ordinary generation.
Given a trajectory prefix $h_t$, the predictive entropy is defined as:
\begin{equation}
\begin{aligned}
H(h_t)
=
-
\sum_{v}
p_\theta(v \mid h_t)
\log p_\theta(v \mid h_t),
\end{aligned}
\end{equation}
where $v$ denotes a token in the vocabulary.
A higher entropy indicates greater uncertainty in the model's subsequent generation.
Motivated by prior findings that tool interactions can produce high-entropy decision points suitable for targeted branching~\citep{arpo, toollight}, we use entropy to prioritize candidate positions for local invoke/skip comparison during pair construction.

\begin{figure}[t!]
    \centering
    \includegraphics[width=\linewidth]{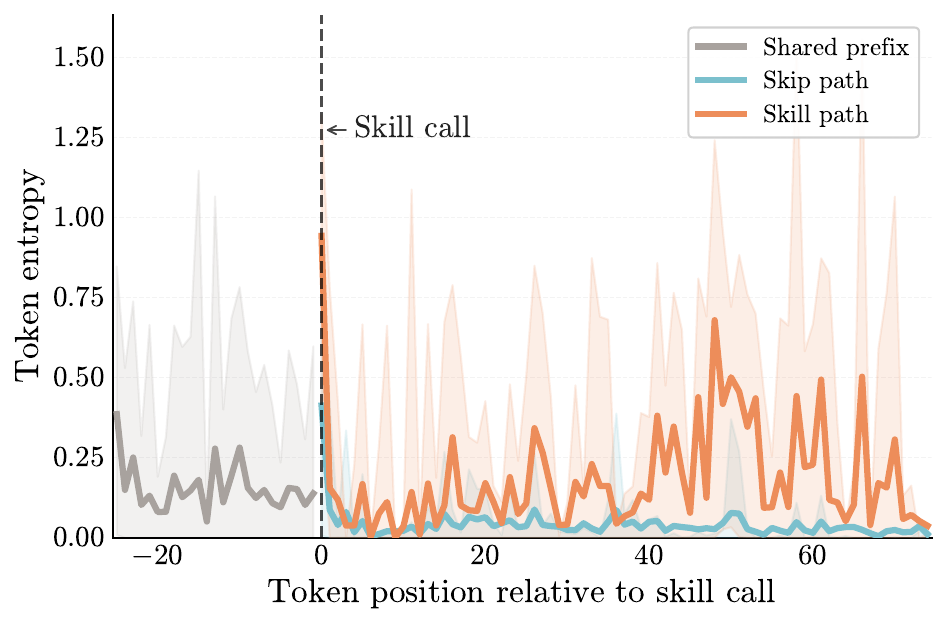}
    \caption{\textbf{Token entropy around invoke/skip points.}}
    \label{fig:entropy_fork_base_left}
\vspace{-12pt}
\end{figure}

We further examine this heuristic through an entropy-fork analysis on ALFWorld~\citep{alfworld}.
As a diagnostic analysis, we create invoke/skip forks at actual skill-call positions and compare the token-level entropy of the two paths after the fork, as shown in Figure~\ref{fig:entropy_fork_base_left}.
The invoke path with skill injection exhibits higher average token entropy in subsequent action prediction.
This suggests that skill injection often increases the model's uncertainty when integrating the returned skill information into the next actions.
We therefore use predictive entropy as a lightweight signal to prioritize candidate branch points that are more likely to produce informative invoke/skip comparisons.

For each selected skill-decision point $(h_t, s)$, we construct two continuations: one that invokes skill $s$ and one that skips it.
Both continuations start from the same prefix $h_t$ and differ only in the forced local decision: invoking skill $s$ or skipping it.
We roll out both branches to the end of the episode and assign labels by an outcome-efficiency rule.
A successful continuation is preferred over a failed one.
If both continuations succeed, we prefer the shorter one, measured by the number of environment steps after the branch.
If both continuations fail, we discard the branch.
This keeps redundant-but-successful skill calls in the training signal and encourages the model to skip unnecessary skills.
\begin{equation}
(c_t^+, c_t^- \mid h_t, s),
\end{equation}
where $c_t^+$ and $c_t^-$ denote the preferred and dispreferred continuations, respectively, under this outcome-efficiency ordering.

This construction differs from episode-level pairing because the two continuations share the same history $h_t$ and are generated by an explicit invoke/skip intervention at the branch point.
Thus, although the label is still evaluated by downstream outcome and efficiency, the comparison controls for the pre-branch trajectory and isolates the immediate skill-or-skip choice more directly than pairing independently sampled complete trajectories.

To focus this signal on the local invoke/skip decision, we compute the DPO loss only within a local window after the branching point.
Specifically, we apply a local loss mask $M_t^{(n)}$ to the continuation and keep only the next $n$ assistant turns after the branch for local DPO.
This makes the gradients more directly target the short-term consequences of the invoke/skip decision rather than the entire continuation of each branch.

Local decision-point preferences concentrate the learning signal around specific invoke/skip decisions, while episode-level preferences provide full-trajectory constraints.

\subsection{Preference Optimization}
\label{subsec:preference_optimization}

We optimize the constructed preference data using Direct Preference Optimization (DPO)~\citep{dpo}.
For a conditioning input $z$ and an output $y$, we define
$r_\theta(z,y)=\log \pi_\theta(y \mid z)-\log \pi_{\mathrm{ref}}(y \mid z)$,
where $\pi_\theta$ is the trainable model and $\pi_{\mathrm{ref}}$ is the reference model.
Given a preference pair $(y^+, y^-)$, the DPO loss is
\begin{equation}
\begin{split}
\mathcal{L}_{\mathrm{DPO}}
= -\mathbb{E}_{(z,y^+,y^-)\sim\mathcal{D}}
\Big[
\log \sigma \Big(
\beta r_\theta(z,y^+)
\\
\quad
-
\beta r_\theta(z,y^-)
\Big)
\Big],
\end{split}
\label{eq:dpo}
\end{equation}
where $\beta$ is the DPO temperature.

We merge episode-level preference pairs and local decision-point preference pairs into the training data and optimize them with the same DPO objective.
For episode-level pairs, $z$ is the task context, while $y^+$ and $y^-$ are the successful and failed trajectories.
For local pairs, the model is conditioned on the pre-branch history $h_t$ and the full skill metadata list, while $s$ only identifies the candidate skill used to create the forced invoke/skip fork.
The paired continuations $(y^+,y^-)$ are then ordered by the outcome-efficiency rule.
During optimization, $M_t^{(n)}$ masks out tokens outside the first $n$ assistant turns after the branch, so local DPO trains only on the selected assistant-generated tokens.

\section{Experiments}
\label{sec:experiments}


\subsection{Experimental Setup}
\label{subsec:exp_setup}

\paragraph{Benchmarks.}
We evaluate SelSkill on four benchmarks.
ALFWorld~\citep{alfworld} evaluates multi-step embodied decision making, and BFCL~\citep{bfcl} evaluates multi-turn function calling.
We further use PopQA~\citep{popqa} and Tau-bench~\citep{tau-bench} to evaluate out-of-domain transfer.

\paragraph{Backbones.}
We use models with basic task-solving ability and valid skill-call formats, allowing SelSkill to focus on selective invocation.
For ALFWorld, since raw models are unstable in environment interaction, we train no-skill Qwen3-4B/8B \cite{qwen3} policies with GRPO~\cite{grpo} as \textbf{RL-Init}, and enable skills afterward.
For BFCL, we use Qwen3-14B as \textbf{Base}, which already supports reliable function-style skill calls through prompting without additional RL initialization.
We follow standard benchmark-specific evaluation protocols and configurations.
Details are provided in Appendix~\ref{app:experimental_details}.

\paragraph{Baselines.}
We compare \textbf{SelSkill} with No-Skill, skill-enabled baselines without selective-invocation training (\textbf{RL-Init w/ Skill} for ALFWorld and \textbf{Base w/ Skill} for BFCL), and signal ablations (Episode only, Entropy-local only, and Skill-call-local only).
For out-of-domain evaluation, we compare No-Skill with +Skill, which enables target-benchmark skills without additional training.
Appendix~\ref{app:selective_baselines} reports additional engineering and adapted selective-invocation baselines.

\paragraph{Metrics.}
We report the following metrics.
\textbf{SR} measures the task success rate.
\textbf{Exec.~Prec.} measures whether a skill invocation is valid and successfully completed.
It is the fraction of attempted invocations that are valid and successfully completed. Invalid, malformed, and execution-error calls count as failures.
For executable skills, this means the skill can interact with the environment without execution errors caused by unmet preconditions, invalid arguments, or malformed calls; for memory skills, this means the call returns valid content for subsequent generation.
SR@Invoke and SR@Skip report success rates conditioned on whether an episode invokes at least one skill.
Their changes provide an indirect view of how selectively the model invokes skills across different trajectory states.
We additionally report Skill/ep as an indicator of invocation frequency and Avg.\ Steps as an indicator of trajectory efficiency.

\begin{table*}[t]
\small
\setlength{\tabcolsep}{5.2pt}
\renewcommand{\arraystretch}{1.2}
\centering
\begin{tabular}{lcccccc}
  \toprule[1.5pt]
  \textbf{Method}
       & \textbf{SR} {(\color{green!40!black}$\uparrow$)}
       & \textbf{Exec.~Prec.} {(\color{green!40!black}$\uparrow$)}
       & \textbf{SR@Invoke} {(\color{green!40!black}$\uparrow$)}
       & \textbf{SR@Skip} {(\color{green!40!black}$\uparrow$)}
       & \textbf{Skill/ep}
       & \textbf{Avg.~Steps} {(\color{green!40!black}$\downarrow$)} \\
  \midrule

  \rowcolor{gray!15}
  \multicolumn{7}{c}{\emph{ALFWorld benchmark}} \\

  Qwen3-4B (No-Skill)
                            &66.4&---&---&---&---&26.2 \\
  \hspace{15pt}+RL-Init w/ Skill
                            &69.5&87.8&81.4&63.5&0.58&23.8 \\

  \rowcolor{blue!3}
  \hspace{15pt}+SelSkill Round1
                            &73.4&81.6&86.8&64.0&0.68&21.5 \\
  \rowcolor{blue!3}
  \hspace{15pt}+SelSkill Round2
                            &\textbf{77.3}&\textbf{96.0}&\textbf{92.6}&\textbf{66.2}&0.59&\textbf{21.3} \\

  Qwen3-8B (No-Skill)
                            &78.9&---&---&---&---&22.3 \\
  \hspace{15pt}+RL-Init w/ Skill
                            &75.8&70.9&73.7&78.8&2.55&24.0 \\

  \rowcolor{blue!3}
  \hspace{15pt}+SelSkill Round1
                            &82.8&94.1&90.7&78.8&0.66&19.7 \\
  \rowcolor{blue!3}
  \hspace{15pt}+SelSkill Round2
                            &85.9&96.6&91.4&\textbf{83.9}&0.46&\textbf{16.3} \\
  \rowcolor{blue!3}
  \hspace{15pt}+SelSkill Round3
                            &\textbf{86.7}&\textbf{100.0}&\textbf{97.0}&83.2&0.44&16.9 \\

  \midrule

  \rowcolor{gray!15}
  \multicolumn{7}{c}{\emph{BFCL benchmark}} \\

  Qwen3-14B (No-Skill)
                            &14.1&---&---&---&---&24.2 \\
  \hspace{15pt}+Base w/ Skill
                            &18.5&44.0&13.4&22.8&0.73&18.2 \\

  \rowcolor{blue!3}
  \hspace{15pt}+SelSkill Round1
                            &22.6&69.7&14.8&30.8&0.92&\textbf{14.1} \\
  \rowcolor{blue!3}
  \hspace{15pt}+SelSkill Round2
                            &\textbf{24.2}&\textbf{73.5}&\textbf{18.2}&\textbf{32.4}&1.01&14.2 \\

  \bottomrule[1.5pt]
\end{tabular}
\caption{\textbf{Performance comparison of different baselines on ALFWorld and BFCL benchmarks.} {\color{blue!30}Blue} denotes our methods. For each backbone, the best results are highlighted in \textbf{bold}.}
\label{tab:main_results}
\vspace{-12pt}
\end{table*}

\subsection{Main Results}
\label{subsec:main_results}

\paragraph{Performance on ALFWorld.}
Table~\ref{tab:main_results} reports the main results on ALFWorld. The experimental results can be summarized in three points:

\textbf{(i) Necessity of selective invocation.}
Simply enabling skills on RL-Init does not lead to consistent performance gains: it improves task success in some settings but degrades it in others.
This inconsistency suggests that the model cannot yet reliably determine when and how to invoke skills.

\textbf{(ii) Improved invocation reliability.}
Compared with skill-enabled initialization, SelSkill substantially improves execution precision and shortens trajectories overall.
For Qwen3-8B, Exec.~Prec. increases from 70.9\% to 100.0\%, while Avg.~Steps decreases from 24.0 to 16.9 by Round3.

\textbf{(iii) A narrow but reliable invocation boundary.}
For Qwen3-8B, SelSkill Round3 achieves 97.0\% SR@Invoke while reducing Skill/ep from 2.55 to 0.44, suggesting that the model learns a more selective and reliable invocation policy.
Qwen3-4B likewise improves SR and SR@Invoke, while keeping its invocation frequency close to the skill-enabled initialization.

\paragraph{Performance on BFCL.}
Table~\ref{tab:main_results} also reports the main results on BFCL. The experimental results can be summarized in three points:

\textbf{(i) Incremental value of skill access.}
Enabling skills improves SR from 14.1\% to 18.5\%, showing that skills can provide additional value in multi-turn function-calling tasks, while leaving room for learning more reliable invocation behavior.

\textbf{(ii) Execution precision as a key bottleneck.}
\textbf{Base w/ Skill} achieves only 44.0\% Exec.~Prec., indicating that many skill calls are not validly and successfully executed.

\textbf{(iii) Higher-quality invocation.}
Compared with \textbf{Base w/ Skill}, SelSkill Round2 improves SR from 18.5\% to 24.2\% and Exec.~Prec. from 44.0\% to 73.5\%, while reducing Avg.~Steps from 18.2 to 14.2.
Notably, Skill/ep increases from 0.73 to 1.01 rather than decreasing.
This suggests that SelSkill does not simply suppress skill use; instead, it enables more reliable invocations, higher task success, and shorter trajectories.
Appendix~\ref{app:selective_baselines} provides a concise BFCL failure breakdown.

\subsection{Ablation Study}
\label{subsec:ablation}

\begin{table}[t!]
\centering
\footnotesize
\setlength{\tabcolsep}{3.3pt}
\renewcommand{\arraystretch}{1.08}
\begin{tabular}{p{0.38\linewidth}ccc}
  \toprule[1.5pt]
  \textbf{Setting}
       & \textbf{SR} {(\color{green!40!black}$\uparrow$)}
       & \textbf{Skill/ep}
       & \textbf{Exec.~Prec.} {(\color{green!40!black}$\uparrow$)} \\
  \midrule

  \rowcolor{gray!15}
  \multicolumn{4}{c}{\emph{Episode-level}} \\
  \makecell[l]{Episode only \\ (standard DPO)} & 75.0 & 1.30 & 75.4 \\
  \midrule
  \rowcolor{gray!15}
  \multicolumn{4}{c}{\emph{Step-level}} \\
  Entropy-local only                &70.3&1.77&81.0 \\
  Skill-call-local only             &80.5&4.29&41.7 \\

  \midrule
  \rowcolor{gray!15}
  \multicolumn{4}{c}{\emph{Episode-level + Step-level}} \\
  \rowcolor{red!3}
  SelSkill ($n=1$)                  &79.7&0.80&91.2 \\
  \rowcolor{red!3}
  SelSkill ($n=3$)                  &\textbf{82.8}&0.66&\textbf{94.1} \\
  \rowcolor{red!3}
  SelSkill ($n=\mathrm{all}$)       &82.0&0.72&92.6 \\

  \bottomrule[1.5pt]
\end{tabular}
\caption{\textbf{Ablation study on ALFWorld benchmark using Qwen3-8B.}
{\color{red!30}Red} denotes mixed-signal training, and \(n\) denotes the number of post-branch assistant turns covered by the loss mask.}
\label{tab:ablation}
\vspace{-12pt}
\end{table}

\begin{table*}[t!]
\centering
\small
\setlength{\tabcolsep}{5.0pt}
\renewcommand{\arraystretch}{1.08}
\begin{tabular}{lccccccc}
  \toprule[1.5pt]
  \multirow{2}{*}{\vspace{-2mm}\textbf{Method}}
       & \multirow{2}{*}{\vspace{-2mm}\textbf{Overall EM} {(\color{green!40!black}$\uparrow$)}}
       & \multicolumn{2}{c}{\textbf{High-pop}}
       & \multicolumn{2}{c}{\textbf{Mid-pop}}
       & \multicolumn{2}{c}{\textbf{Low-pop}} \\
  \cmidrule(lr){3-4}
  \cmidrule(lr){5-6}
  \cmidrule(lr){7-8}
       &
       & \textbf{EM} {(\color{green!40!black}$\uparrow$)} & \textbf{Skill Rate}
       & \textbf{EM} {(\color{green!40!black}$\uparrow$)} & \textbf{Skill Rate}
       & \textbf{EM} {(\color{green!40!black}$\uparrow$)} & \textbf{Skill Rate} \\
  \midrule
  Base
       &20.1&42.2&---&8.4&---&9.6&--- \\
  \hspace{3pt}+Skill
       &61.0&58.4&78\%&62.0&93\%&62.7&94\% \\
  \rowcolor{red!3}
  SelSkill
       &\textbf{62.9}&\textbf{60.2}&57\%&\textbf{63.9}&87\%&\textbf{64.5}&87\% \\
  \bottomrule[1.5pt]
\end{tabular}
\caption{\textbf{OOD transfer on PopQA benchmark.}}
\label{tab:ood_popqa}
\vspace{-10pt}
\end{table*}

We conduct ablation experiments on ALFWorld with Qwen3-8B to examine the contributions of different preference signals.
All variants use the same one-round training setting as \textbf{SelSkill Round1}.
SelSkill combines episode-level preferences, which compare successful and failed trajectories, with entropy-guided step-level preferences, which supervise local invoke/skip decisions.
The loss-mask parameter \(n\) controls how many post-branch assistant turns are included in the local training objective.
Table~\ref{tab:ablation} reports the results, which can be summarized in three points:

\textbf{(i) Episode-level supervision is insufficient.}
\textbf{Episode only}, which corresponds to standard DPO using only episode-level preference pairs, achieves 75.0\% SR and 75.4\% Exec.~Prec.
This suggests that trajectory-level preference learning alone cannot reliably supervise local skill invocation decisions.

\textbf{(ii) Local supervision alone is unbalanced.}
\textbf{Entropy-local only} achieves relatively high Exec.~Prec. but low SR, while \textbf{Skill-call-local only} improves SR to 80.5\% but reduces Exec.~Prec. to 41.7\% with 4.29 Skill/ep.
This indicates that local supervision alone may improve one aspect of invocation behavior while sacrificing others.

\textbf{(iii) Mixed signals achieve the best balance.}
\textbf{SelSkill} with \(n=3\) achieves the highest SR of 82.8\% and Exec.~Prec. of 94.1\%, with 0.66 Skill/ep.
These results show that combining episode-level and step-level preferences improves both task success and invocation quality, rather than relying on more frequent skill calls.

\subsection{Out-of-Domain Generalization}
\label{subsec:ood}

A key question is whether the selective invocation ability learned on BFCL can generalize to new domains with previously unseen skills.
We evaluate BFCL SelSkill Round2 in a zero-shot manner on two OOD benchmarks that are not seen during training.
PopQA uses Wikipedia retrieval skills for knowledge-intensive question answering, while Tau-bench evaluates multi-turn service-oriented agent tasks with benchmark-specific skills and a GPT-4.1 user simulator.
Tables~\ref{tab:ood_popqa} and~\ref{tab:ood_taubench} report the results on PopQA and Tau-bench, respectively.
The results can be summarized in three points:

\textbf{(i) Partial transfer across domains and unseen skills.}
Neither benchmark uses the BFCL skills available during training, so this evaluation tests the transfer of \emph{invocation judgment} rather than memorization of specific skill knowledge.
On both OOD benchmarks, enabling benchmark-specific skills improves performance over the no-skill baseline, showing that the newly provided skills are useful in their target domains.
SelSkill further improves over +Skill on both benchmarks, suggesting partial transfer of its invocation criterion.

\textbf{(ii) Selective retrieval on PopQA.}
In this setting, Skill Rate denotes the percentage of examples where the model invokes a Wikipedia retrieval skill. Since PopQA is single-turn, Skill Rate is equivalent to \textbf{Skill/ep}.
We use entity popularity as a rough proxy for how likely the answer is to be covered by the model's parametric knowledge: high-popularity entities are more likely to be internalized by the model, while mid- and low-popularity entities typically require external retrieval more often.
Table~\ref{tab:ood_popqa} shows that SelSkill improves EM across all popularity groups while reducing Skill Rate in an intuitive way.
For high-popularity entities, Skill Rate drops substantially from 78\% to 57\%, while EM improves from 58.4 to 60.2.
In contrast, for mid- and low-popularity entities, SelSkill only slightly reduces Skill Rate, from 93\% and 94\% to 87\%, while improving EM from 62.0/62.7 to 63.9/64.5.
This indicates that SelSkill does not simply suppress skill use; instead, it tends to preserve retrieval for examples that likely require external knowledge and skip unhelpful calls when parametric knowledge is more likely to suffice.

\begin{table}[t!]
\centering
\small
\setlength{\tabcolsep}{3.5pt}
\renewcommand{\arraystretch}{1.10}
\vspace{5pt}
\begin{tabular}{lcccc}
  \toprule[1.5pt]
  \textbf{Method}
       & \textbf{Avg.} {(\color{green!40!black}$\uparrow$)}
       & \textbf{Air.} {(\color{green!40!black}$\uparrow$)}
       & \textbf{Ret.} {(\color{green!40!black}$\uparrow$)}
       & \textbf{SR@Inv./Skip} {(\color{green!40!black}$\uparrow$)} \\
  \midrule
  Base
       &31.9&22.0&41.7&--- / --- \\
  \hspace{3pt}+Skill
       &39.6&\textbf{34.0}&45.2&41.7 / 47.8 \\
  \rowcolor{red!3}
  SelSkill
       &\textbf{41.4}&\textbf{34.0}&\textbf{48.7}&\textbf{50.0} / \textbf{49.2} \\
  \bottomrule[1.5pt]
\end{tabular}
\caption{\textbf{OOD transfer on Tau-bench.}
Avg. is mean pass@1 over airline (Air.) and retail (Ret.) domains.}
\label{tab:ood_taubench}
\vspace{-18pt}
\end{table}

\textbf{(iii) More reliable invocation on Tau-bench.}
On Tau-bench, we evaluate SelSkill zero-shot with unseen APIs, policies, and skills.
Compared with +Skill, it improves average pass@1 from 39.6 to 41.4.
Both Avg. and SR@Invoke are macro-averaged across airline and retail; the latter increases from 41.7 to 50.0, indicating that invoked episodes succeed more often in new settings.






\section{Analyses}
\label{sec:analyses}

\paragraph{Gradient localization.}
\label{subsec:gradient_focus}

\begin{figure}[t!]
\centering
\includegraphics[width=0.95\linewidth]{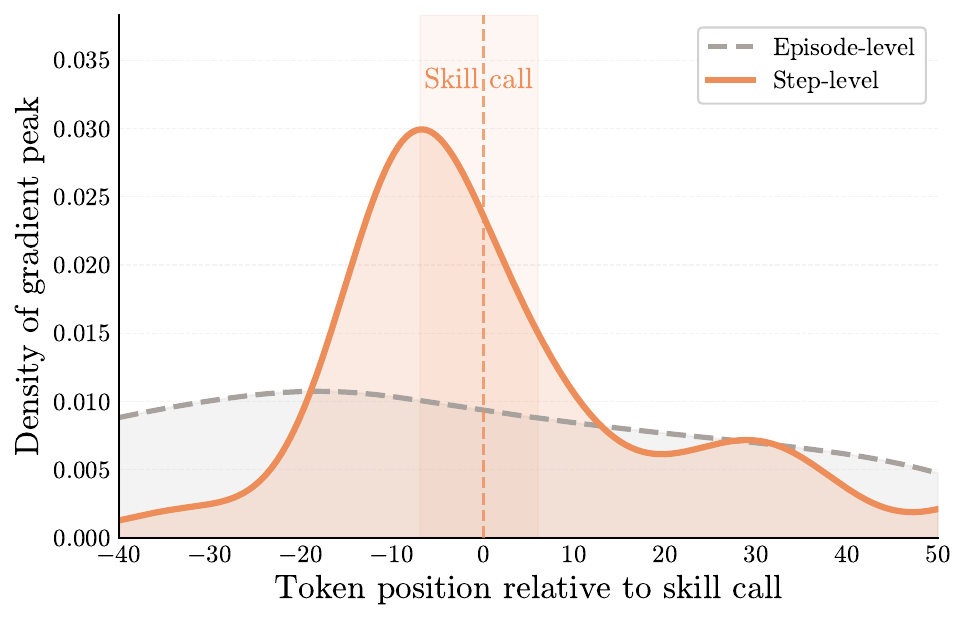}
\caption{
\textbf{Step-level preferences produce more localized gradient peaks around the skill-call token.}
}
\label{fig:gradient_focus}
\vspace{-12pt}
\end{figure}

To examine whether step-level preferences provide more localized supervision for skill invocation, we compare gradient peak positions between ALFWorld episode-level pairs from complete trajectories and entropy-guided step-level pairs branched at high-uncertainty skill-decision points.
For each pair, we align the skill-call position to 0, record the token position with the largest gradient norm, and visualize its distribution using kernel density estimation.
The vertical axis in Figure~\ref{fig:gradient_focus} represents the estimated density of gradient peak positions.

Figure~\ref{fig:gradient_focus} shows that episode-level preferences produce more dispersed gradient peaks, whereas step-level preferences concentrate them around the skill-call region.
This suggests that episode-level preferences provide broad trajectory-level guidance, while step-level preferences more directly supervise local invoke/skip decisions.
The two signals thus provide complementary supervision for selective skill invocation, consistent with the advantage of mixed-signal training in Table~\ref{tab:ablation}.

\paragraph{Additional analyses.}
Appendix~\ref{app:counterfactual_benefit} uses counterfactual comparisons to examine when skill calls are useful.
Skill invocation improves only a small fraction of trajectories, and harmful calls can still be semantically plausible.
Appendix~\ref{app:rl_selectivity} examines episode-level reinforcement learning and finds that it does not reliably calibrate local invocation decisions.
Skill usage remains unstable even when overall task success improves.
Appendix~\ref{app:selective_baselines} compares SelSkill with engineering and adapted selective-invocation baselines.
These simple alternatives do not replace selectivity training.
Appendix~\ref{app:entropy_branching} evaluates entropy-guided branching.
It selects more informative local comparisons than random branching and achieves comparable performance to all-skill branching at lower sampling cost.
Appendix~\ref{app:robustness} tests robustness under injected distractor skills.
SelSkill almost never invokes noise skills and shows only modest degradation with the largest skill listing.
Finally, Appendix~\ref{app:trajectory_cases} presents cases in which skills are unnecessary, invoked before their preconditions are met, or beneficial when invoked at the appropriate time.

\section{Conclusion}
\label{sec:conclusion}

We formulate selective skill invocation as a \emph{skill-or-skip} decision and propose \textbf{SelSkill}, which learns selective invocation policies from combined episode- and step-level preferences.
On ALFWorld, SelSkill improves task success by \textbf{10.9} points over the skill-enabled baseline and \textbf{7.8} points over No-Skill, while improving execution precision by \textbf{29.1} points. On BFCL, task success and execution precision improve by \textbf{5.7} and \textbf{29.5} points, respectively.
Zero-shot results on PopQA and Tau-bench suggest partial transfer to unseen domains and skills.


\section*{Limitations}

This work studies selective skill invocation with a fixed, offline-constructed skill library, while its construction and evolution are beyond our scope. We use predictive entropy as a lightweight heuristic for prioritizing candidate decision points, leaving alternative criteria to future work. Our evaluation is limited to Qwen3 backbones. Finally, our benchmarks do not fully capture deployments with irreversible external effects.


\bibliography{custom}

\clearpage
\appendix
\addtocontents{toc}{\protect\appendixTOCstart}
\renewcommand{\contentsname}{Appendix}

\begingroup
\setcounter{tocdepth}{3}
\makeatletter

\let\oldl@subsection\l@subsection
\renewcommand{\l@subsection}[2]{%
  \vspace{0.6ex}
  \oldl@subsection{#1}{#2}%
}

\let\oldl@subsubsection\l@subsubsection
\renewcommand{\l@subsubsection}[2]{%
  \vspace{0.6ex}
  \oldl@subsubsection{#1}{#2}%
}

\makeatother
\tableofcontents
\endgroup

\appendix
\section{Skill Library}
\label{app:skill_library}

We study selective skill invocation under a fixed skill library.
All skills are constructed offline before policy optimization, using only training resources such as training trajectories, task instructions, API schemas, and environment documentation.
These skills remain unchanged throughout SelSkill training and evaluation, and no evaluation examples are used during skill construction.

For each benchmark, we identify recurring procedural patterns from training resources and consolidate them into reusable callable skills.
Each skill consists of lightweight metadata and full skill content.
The metadata is visible to the model before invocation and supports the skill-or-skip decision, while the full content is injected into the context or executed only after the model explicitly chooses to invoke the skill.

Skill construction follows two principles.
First, each skill should capture a reusable procedure that can apply across multiple task instances, such as checking preconditions, following environment constraints, retrieving supporting evidence, or executing a common action sequence.
It should not encode an instance-specific solution or a shortcut tailored to a particular example.
Second, skill content should avoid any evaluation leakage.
It must not contain evaluation answers, evaluation trajectories, target states, case identifiers, or other information that would allow the model to solve an evaluation instance by memorization rather than by deciding when to invoke a fixed skill.

Table~\ref{tab:skill_examples} shows representative skills from each benchmark.
The entries in the table are summarized and shortened; full skill definitions are provided in the supplementary materials.

\begin{table*}[t!]
\centering
\small
\setlength{\tabcolsep}{4.5pt}
\renewcommand{\arraystretch}{1.16}
\begin{tabular}{
@{}p{0.15\linewidth}
c
p{0.15\linewidth}
p{0.255\linewidth}
p{0.265\linewidth}@{}}
\toprule[1.5pt]
\textbf{Benchmark}
& \textbf{Skills}
& \textbf{Example}
& \textbf{Before invocation}
& \textbf{After invocation} \\
\midrule

\textbf{ALFWorld}
& 10 (4 / 6)
& heat object
& Use when a task requires a heated object.
& Check object and microwave availability, then perform the heating procedure. \\

\addlinespace
\textbf{BFCL}
& 18 (10 / 8)
& place stock order
& Use for stock order placement or verification.
& Check ticker, side, quantity, order type, and price, while preserving user-specified limit prices. \\

\addlinespace
\textbf{Tau-bench airline}
& 11 (4 / 7)
& cancel flight
& Use when a user requests flight cancellation.
& Check cancellation eligibility under airline policy before calling the cancellation API. \\

\addlinespace
\textbf{Tau-bench retail}
& 6 (2 / 4)
& cancel order
& Use when a user requests order cancellation.
& Check order status, cancellation window, and item eligibility before calling the cancellation API. \\

\addlinespace
\textbf{PopQA}
& 4 (4 / 0)
& lookup person fact
& Use when uncertain about a person's biographical attribute.
& Retrieve Wikipedia evidence and extract the requested attribute. \\

\bottomrule[1.5pt]
\end{tabular}
\caption{
\textbf{Representative skills from each benchmark.}
Skill counts are reported as total (memory / executable).
Entries are summarized and shortened.
}
\label{tab:skill_examples}
\vspace{-6pt}
\end{table*}

\section{Counterfactual Skill Benefit}
\label{app:counterfactual_benefit}

We conduct counterfactual experiments to characterize when skill access actually changes the execution trajectory.
For each benchmark and model, we deliberately construct a counterfactual pair for the same task instance: a skill-disabled trajectory and a skill-enabled trajectory under the same task setting.
We then align the two runs by case ID and compare whether enabling skills changes the final outcome.
This design compares outcome changes under skill access on the same case,
rather than comparing different tasks or different sampled instances.

The analysis covers 513 paired runs across multiple benchmarks and model families:
BFCL and Tau-bench with Qwen3-14B, ALFWorld with Qwen3-8B, and BFCL with Gemini-2.5-flash-lite.
For BFCL, we use the base and long-context multi-turn splits and exclude missing-function and missing-parameter categories, where the required function or parameter information is unavailable and skill use is therefore structurally blocked.
For Tau-bench and ALFWorld, we align paired runs by task ID/trial and game file, respectively.

\begin{table*}[t!]
\centering
\small
\setlength{\tabcolsep}{4.5pt}
\renewcommand{\arraystretch}{1.12}
\vspace{5pt}

\begin{tabular}{lccccc}
\toprule[1.5pt]
\multicolumn{6}{c}{\textbf{Panel A: Counterfactual outcome changes}} \\
\midrule
\multirow{2}{*}{Outcome}
& BFCL Qwen3
& ALFWorld Qwen3
& Tau-bench Qwen3
& BFCL Gemini
& Total \\
& $(n=124)$ & $(n=84)$ & $(n=181)$ & $(n=124)$ & $(n=513)$ \\
\midrule
Positive
& 21 (16.9\%) & 9 (10.7\%) & 34 (18.8\%) & 5 (4.0\%) & \textbf{69 (13.5\%)} \\
Negative
& 6 (4.8\%) & 1 (1.2\%) & 20 (11.0\%) & 16 (12.9\%) & \textbf{43 (8.4\%)} \\
No clear effect
& 97 (78.2\%) & 74 (88.1\%) & 127 (70.2\%) & 103 (83.1\%) & \textbf{401 (78.2\%)} \\
\bottomrule[1.5pt]
\end{tabular}

\vspace{0.8em}

\begin{tabular}{lccc}
\toprule[1.5pt]
\multicolumn{4}{c}{\textbf{Panel B: Semantic relevance of invoked skills}} \\
\midrule
Outcome
& BM25
& Embedding cosine (E5)
& Interpretation \\
\midrule
Positive
& 4.58
& 0.721
& Relatively relevant \\
Negative
& 4.19
& 0.719
& Relatively relevant \\
No clear effect
& 3.62
& 0.703
& Relatively less relevant \\
\bottomrule[1.5pt]
\end{tabular}

\caption{
\textbf{Counterfactual analysis of skill benefit.}
\textbf{Panel A} reports outcome changes after enabling skills.
\textbf{Panel B} reports similarity between task instructions and invoked skill metadata, showing that both helpful and harmful invocations can appear semantically relevant.
}
\label{tab:counterfactual_skill_benefit}
\vspace{-12pt}
\end{table*}

Table~\ref{tab:counterfactual_skill_benefit} shows that skill benefits are highly concentrated.
Across 513 paired runs, enabling skills improves the trajectory in only 13.5\% of cases, harms an otherwise correct trajectory in 8.4\%, and has no clear effect in 78.2\%.
Thus, skill access is not uniformly beneficial: most cases either do not require the skill or cannot be changed by it, while a smaller but non-negligible set of cases is sensitive to the invocation decision.

Panel B further examines whether harmful invocations are simply caused by choosing semantically unrelated skills.
We compute BM25 and embedding-based cosine similarity between the task instruction and the invoked skill metadata, using E5 \cite{e5} as the embedding model.
Negative cases have similarity scores close to Positive cases and higher than No-clear-effect, indicating that harmful invocations are often semantically plausible.
The BM25 gap between effective or harmful invocations and unnecessary invocations is significant ($p=0.042$), while embedding similarity shows a consistent but non-significant trend.
This suggests that many failures arise after the model has already selected a plausible skill: the harder problem is deciding whether the current state provides the right conditions and timing for invoking it effectively.

\begin{table*}[t!]
\centering
\small
\setlength{\tabcolsep}{4.5pt}
\renewcommand{\arraystretch}{1.08}
\begin{tabular}{c|ccc|ccc}
\toprule[1.5pt]
\textbf{Step}
& \textbf{w/o Skill SR} {(\color{green!40!black}$\uparrow$)}
& \textbf{w/ Skill SR}{(\color{green!40!black}$\uparrow$)}
& \textbf{$\Delta$}
& \textbf{Skill/ep}
& \textbf{SR@Invoke}{(\color{green!40!black}$\uparrow$)}
& \textbf{SR@Skip} {(\color{green!40!black}$\uparrow$)}\\
\midrule
0  & 19.5 & 21.4 & +1.9  & 11.13 & 20.7 & 50.0 \\
5  & 24.2 & 19.0 & -5.2  & 1.25  & 24.3 & 14.3 \\
10 & 26.6 & 39.1 & +12.5 & 3.26  & 30.8 & 63.6 \\
15 & 28.9 & 45.8 & +16.9 & 12.25 & 33.8 & 100.0 \\
20 & 42.2 & 54.0 & +11.8 & 1.84  & 13.5 & 84.0 \\
25 & 44.5 & 66.3 & +21.8 & 2.84  & 31.7 & 100.0 \\
30 & 52.3 & 48.3 & -4.0  & 0.97  & 9.8  & 82.6 \\
35 & 63.3 & 63.9 & +0.6  & 3.18  & 19.4 & 97.9 \\
40 & 79.7 & 59.3 & -20.4 & 2.45  & 22.2 & 100.0 \\
\bottomrule[1.5pt]
\end{tabular}
\caption{
\textbf{Comparison of GRPO+KL Performance on ALFWorld benchmark}.
All SR values are percentages.
SR@Invoke and SR@Skip are computed on validation episodes with and without skill calls, respectively.
}
\label{tab:online_rl_selectivity}
\vspace{-12pt}
\end{table*}

\section{Episode-Level RL Does Not Reliably Calibrate Skill Use}
\label{app:rl_selectivity}

We compare two GRPO+KL variants on ALFWorld to examine whether episode-level reward can learn selective skill invocation.
Both variants start from Qwen3-8B base and use the same training hyperparameters:
learning rate $1\mathrm{e}{-6}$, group size 8, train batch size 32, mini-batch size 64, KL coefficient 0.01, maximum 50 steps per episode, and validation temperature 0.4.
\textsc{GRPO-w/oSkill} does not receive any skill listing and is trained with task reward only: $+10$ for success and $-0.1$ for invalid actions.
\textsc{GRPO-w/Skill} receives the skill listing at every step and uses the same task reward plus a skill-success bonus of $+1.0$ whenever a skill executes without error.

Table~\ref{tab:online_rl_selectivity} shows that episode-level RL improves task success at some checkpoints, but does not produce stable skill-use behavior.
The advantage of \textsc{GRPO-w/Skill} over \textsc{GRPO-w/oSkill} fluctuates across training, and skill calls per episode do not converge to a consistent pattern.
Episodes without skill calls often achieve higher success rates than episodes with skill calls, suggesting that the learned policy still invokes skills in many low-yield states. This does not mean that skill calls directly cause failure, since invoked episodes may be harder or more uncertain.
Rather, the result shows that sparse task reward, even with a simple skill-success bonus, is too coarse to supervise the local skill-or-skip decision.
The policy can improve final task success while still failing to learn when a skill should be skipped.

Table~\ref{tab:case-f4} illustrates this issue with a representative ALFWorld trajectory.
The model eventually succeeds, but it invokes \texttt{heat\_object} for a task that only requires placing a tomato on the microwave.
The redundant calls waste steps, introduce precondition errors, and make the trajectory much longer than necessary.
Thus, the final success reward cannot distinguish a skill that helped from a skill that the model succeeded despite.
This is the credit assignment problem at the individual skill-call level.
\begin{table*}[t]
\centering
\small
\setlength{\tabcolsep}{6pt}
\renewcommand{\arraystretch}{1.12}
\begin{tabular}{@{}p{0.19\textwidth}p{0.76\textwidth}@{}}
\toprule[1.5pt]
\multicolumn{2}{@{}l}{\textbf{RL-Init model: successful trajectory with redundant skill calls}} \\
\multicolumn{2}{@{}l}{\emph{Task: Put some tomato on microwave.} 
Success in 18 steps; optimal $\approx$6.} \\
\midrule

\textbf{Correct prefix}
&
\texttt{go to fridge 1} $\rightarrow$ 
\texttt{open fridge 1} $\rightarrow$
\texttt{take tomato 3 from fridge 1}.
The agent correctly obtains the tomato. \\

\addlinespace[3pt]
\textbf{Redundant invocation}
&
\texttt{Skill(heat\_object)} is invoked and executes, although the task only requires placing the tomato in the microwave. \\

\addlinespace[3pt]
\textbf{State-breaking action}
&
\texttt{go to countertop 1} $\rightarrow$
\texttt{move tomato 3 to countertop 1}.
The unnecessary skill invocation changes the subsequent behavior, and the agent puts down the tomato. \\

\addlinespace[3pt]
\textbf{Failed invocations}
&
The model invokes \texttt{Skill(heat\_object)} twice more.
Both calls fail because the agent is no longer holding an object, violating the skill precondition. \\

\addlinespace[3pt]
\textbf{Recovery}
&
\texttt{go to countertop 1} $\rightarrow$
\texttt{take tomato 3 from countertop 1} $\rightarrow$
\texttt{Skill(heat\_object)} $\rightarrow$
\texttt{move tomato 3 to microwave 1}.
The task is eventually completed, but only after several unnecessary steps. \\

\bottomrule[1.5pt]
\end{tabular}
\caption{\textbf{Success Despite Redundant Skill Calls.}
An ALFWorld trajectory where the model succeeds on a placement-only task despite unnecessary \texttt{heat\_object} calls that introduce precondition errors and extra steps.}
\label{tab:case-f4}
\vspace{-6pt}
\end{table*}

\section{Additional Selective-Invocation Baselines}
\label{app:selective_baselines}

\paragraph{Engineering baselines.}
We evaluate whether simple engineering changes can replace selectivity training on ALFWorld with Qwen3-8B RL-Init.
\emph{Skill-as-Context} injects all skill content directly into the prompt, \emph{Conservative Prompt} explicitly discourages unnecessary calls, and \emph{Explicit Skip Option} adds a universal skip skill.
None improves over RL-Init w/ Skill, while SelSkill reaches substantially higher SR with fewer and more reliable invocations.

\paragraph{Adapted selective-invocation baselines.}
We adapt two prior selective-use ideas to our setting rather than treat them as direct reproductions.
\emph{Raw-Logit Gate}~\citep{xu-etal-2025-alignment} skips skill access when the no-skill action has low mean token NLL, while \emph{Call-or-Skip Gate}~\citep{when2call} trains a same-backbone SFT invoke/skip classifier using SelSkill's outcome-efficiency labels.
Both improve execution precision but trail SelSkill in SR and SR@Invoke at similar Skill/ep.

\paragraph{BFCL failure analysis.}
Among the remaining attributable BFCL failures, 7.2\% arise from incorrect skill/function decisions, 40.3\% from missing API/environment knowledge, and 52.5\% from downstream reasoning errors.

\begin{table*}[t!]
\centering
\small
\setlength{\tabcolsep}{5.0pt}
\renewcommand{\arraystretch}{1.08}
\begin{tabular}{lccccc}
\toprule[1.5pt]
\textbf{Method}
& \textbf{SR} {(\color{green!40!black}$\uparrow$)}
& \textbf{Exec.~Prec.} {(\color{green!40!black}$\uparrow$)}
& \textbf{Skill/ep}
& \textbf{SR@Invoke} {(\color{green!40!black}$\uparrow$)}
& \textbf{SR@Skip} {(\color{green!40!black}$\uparrow$)} \\
\midrule
No-Skill & 78.9 & --- & 0.00 & --- & 78.9 \\
RL-Init w/ Skill & 75.8 & 70.9 & 2.55 & 73.7 & 78.8 \\
\midrule
Skill-as-Context & 71.9 & --- & --- & --- & --- \\
Conservative Prompt & 71.9 & --- & 1.41 & 78.0 & 66.7 \\
Explicit Skip Option & 70.3 & --- & 2.34 & 72.4 & 67.3 \\
\midrule
Raw-Logit Gate & 77.3 & 95.9 & 0.57 & 68.2 & 82.1 \\
Call-or-Skip Gate & 79.7 & 95.0 & 0.55 & 69.0 & 82.5 \\
\midrule
\rowcolor{red!3}
SelSkill Round1 & 82.8 & 94.1 & 0.66 & 90.7 & 78.8 \\
\rowcolor{red!3}
SelSkill Round2 & 85.9 & 96.6 & 0.46 & 91.4 & \textbf{83.9} \\
\rowcolor{red!3}
SelSkill Round3 & \textbf{86.7} & \textbf{100.0} & \textbf{0.44} & \textbf{97.0} & 83.2 \\
\bottomrule[1.5pt]
\end{tabular}
\caption{\textbf{Engineering and adapted selective-invocation baselines on ALFWorld with Qwen3-8B.}
Raw-Logit Gate and Call-or-Skip Gate are adaptations of prior selective-use formulations rather than direct reproductions.}
\label{tab:engineering_baselines}
\label{tab:adapted_selective_baselines}
\vspace{-8pt}
\end{table*}

\section{Ablation on Entropy-Guided Branching}
\label{app:entropy_branching}

We ablate the branching-point selection strategy used for local decision-point preference construction.
All conditions use the same base model, Qwen3-8B RL-Init on ALFWorld, and the same $K=4$ free-sampling procedure.
To isolate the effect of branching-point selection, we match the final amount of training data to the entropy-guided setting across all strategies.

\begin{table}[t!]
\centering
\footnotesize
\setlength{\tabcolsep}{4.0pt}
\renewcommand{\arraystretch}{1.08}
\begin{tabular}{@{}lcc@{}}
\toprule[1.5pt]
\textbf{Strategy} & \textbf{Pairs/game}{(\color{green!40!black}$\uparrow$)} & \textbf{Inv.-pref. (\%)} \\
\midrule
All-skill      & 0.18 & 31.0 \\
Random         & 0.68 & 15.7 \\
Entropy        & 0.57 & 24.1 \\
\bottomrule[1.5pt]
\end{tabular}
\caption{\textbf{Quality of retained local preference pairs before size matching.}}
\label{tab:entropy_pair_quality}
\vspace{-13pt}
\end{table}

Table~\ref{tab:entropy_pair_quality} reports the retained local preference pairs before final size matching.
Random branching retains many pairs, but only a relatively small portion of them prefer the invoke continuation.
Entropy-guided branching retains a comparable number of pairs while yielding a higher invoke-preferred ratio.
All-skill branching has the highest invoke-preferred ratio, but it retains substantially fewer pairs per game.
This is because it attempts forks at all skill-invocation positions, while many of these positions lead the invoke and skip continuations to the same final outcome and therefore cannot form a clear outcome-efficiency preference.
Thus, trying more branch points does not necessarily produce more valid preference pairs.
These results suggest that entropy-guided selection can more efficiently identify positions where valid local invoke/skip comparisons can be constructed.

\begin{table}[t!]
\centering
\small
\setlength{\tabcolsep}{4.5pt}
\renewcommand{\arraystretch}{1.08}
\begin{tabular}{lccc}
\toprule[1.5pt]
\textbf{Condition} & \textbf{SR} {(\color{green!40!black}$\uparrow$)} & \textbf{Skill/ep} & \textbf{Exec. Prec.} {(\color{green!40!black}$\uparrow$)}\\
\midrule
RL-Init w/ Skill        & 75.8 & 2.55 & 70.9 \\
All-skill        & \textbf{82.8} & 0.95 & 77.0 \\
Random           & 80.5 & 0.59 & \textbf{96.1} \\
Entropy-guided   & \textbf{82.8} & 0.66 & 94.1 \\
\bottomrule[1.5pt]
\end{tabular}
\caption{
\textbf{Experimental Results on ALFWorld benchmark with Different Branching Strategies.}
All values except Skill/ep are percentages.
}
\label{tab:entropy_training_results}
\end{table}

Table~\ref{tab:entropy_training_results} further compares the downstream results after size-matched training.
Entropy-guided branching achieves the same SR as all-skill branching, while the latter requires roughly three times the sampling cost.
It also yields slightly higher SR than random branching.
This indicates that entropy is not an exact causal criterion, but it provides useful guidance for branching-point selection, achieving training performance close to exhaustive branching with much lower sampling cost.

\section{Robustness to Distractor Skills}
\label{app:robustness}

We evaluate whether the invocation decisions of SelSkill Round2 remain robust as the skill listing expands.
Starting from the standard 18-skill BFCL setting, we inject noise skills to create listings of 28, 38, and 68 skills.
The noise skills are synthetically constructed to cover unrelated domains, such as calendar management, music streaming, e-commerce, and fitness tracking, with realistic \texttt{when-to-use} conditions that do not overlap with any BFCL evaluation task.
They provide no task-relevant information and serve only as listing distractors.
All other conditions remain identical to the main experiment.

\begin{table}[t]
\centering
\small
\setlength{\tabcolsep}{6pt}
\begin{tabular}{lcc}
\toprule[1.5pt]
\textbf{Skill listing size} & \textbf{SR}{(\color{green!40!black}$\uparrow$)} & \textbf{Noise skill calls} \\
\midrule
18 skills (standard)   & 24.2 & --- \\
28 skills (+10 noise)  & 24.2 & 0\,/\,301\; (0.0\%) \\
38 skills (+20 noise)  & 25.4 & 5\,/\,250\; (2.0\%) \\
68 skills (+50 noise)  & 21.4 & 0\,/\,263\; (0.0\%) \\
\bottomrule[1.5pt]
\end{tabular}
\caption{\textbf{Robustness of SelSkill Round2 under expanded skill listings.}}
\label{tab:robustness}
\vspace{-12pt}
\end{table}

Table~\ref{tab:robustness} shows that SR remains stable when the skill listing expands from 18 to 38 skills, and only drops modestly when the listing is further expanded to 68 skills.
This decline is likely due to additional context noise from the longer skill listing, rather than incorrect invocations of the injected noise skills.
The model almost never invokes these noise skills, indicating strong robustness to irrelevant or low-quality skills.
This behavior is consistent with the skill-or-skip formulation: the model learns not only to identify potentially relevant skills, but also to skip skills that should not intervene in the current task state.
These results suggest that SelSkill Round2 remains applicable under larger and noisier skill libraries.

\begin{table}[t!]
\centering
\small
\setlength{\tabcolsep}{3.5pt}
\renewcommand{\arraystretch}{1.08}
\begin{tabular}{lll}
\toprule[1.5pt]
\textbf{Setting} & \textbf{ALFWorld} & \textbf{BFCL} \\
\midrule
Backbone & Qwen3-8B/4B & Qwen3-14B \\
Mode & Thinking-style & Non-thinking \\
Initialization & RL-Init & Base \\
Eval decoding & Greedy & Greedy \\
Eval metric & Task success & Exact match \\
SelSkill rounds & 3 (8B) / 2 (4B) & 2 \\
Learning rate & $1{\times}10^{-6}$ & $5{\times}10^{-6}$ \\
Max length & 4096 & 12288 \\
\midrule
Tuning & \multicolumn{2}{l}{Full-parameter fine-tuning} \\
$\beta$ & \multicolumn{2}{l}{0.1} \\
Optimizer & \multicolumn{2}{l}{AdamW with cosine schedule} \\
Warmup & \multicolumn{2}{l}{0.1} \\
Epochs & \multicolumn{2}{l}{3} \\
Local mask & \multicolumn{2}{l}{$n=3$ post-branch assistant turns} \\
\bottomrule[1.5pt]
\end{tabular}
\caption{\textbf{Model, evaluation, and training settings.}}
\label{tab:app_experimental_settings}
\vspace{-12pt}
\end{table}

\section{Experimental Details}
\label{app:experimental_details}

\subsection{Benchmark-Specific Setup}

\paragraph{ALFWorld.}
We keep the benchmark's thinking-style interaction format, because the agent needs to reason over environment observations before producing executable actions.
After RL initialization, skills are enabled through the system prompt, which contains skill metadata and few-shot skill-call examples.
The full skill body is loaded only after the model explicitly invokes the corresponding skill.
Evaluation uses greedy decoding on a fixed held-out split.

\paragraph{BFCL.}
We use Qwen3-14B in non-thinking mode for BFCL.
BFCL contains long multi-turn function-calling conversations, and enabling thinking substantially increases the context length during rollout collection and preference training.
We therefore use non-thinking mode to keep the interaction within the context budget.
Skills are enabled by adding skill metadata and few-shot examples to the system prompt.
The skill metadata follows the BFCL tool-calling format and includes the skill-use condition in the description.
Evaluation uses exact scoring with greedy decoding.

\subsection{Preference Pair Collection}

\paragraph{Episode-level collection.}
For episode-level data, we sample $K=10$ complete trajectories or task outputs for the same input and label them using the final benchmark outcome.
When constructing episode-level pairs, we remove malformed positive samples so that the chosen side does not contain invalid action or call formats.

\paragraph{Local branching collection.}
For local data, we first run the current policy and record token-level
log probabilities during generation. For each sampled instance, we form
two candidate pools: the top-3 high-entropy positions following skill
invocations and the top-3 high-entropy positions during ordinary generation.
The former captures uncertain states after skill intervention, while the
latter captures uncertain states in ordinary execution.
For each selected branch point, we use interrupted rollout with $K = 4$.
Specifically, we use a temporary intervention prompt only during data
collection to elicit invoke and skip continuations from the same trajectory
prefix at that branch point.
Apart from this local collection prompt, all continuations use the same
original prompt, skill listing, decoding setting, and evaluation protocol.
After the local invoke/skip choice is made, generation continues normally
under the current policy. During training, we restore the original prompt
so that the chosen and rejected continuations are conditioned on the same
original prefix.

The resulting continuations are then filtered and ordered by the
outcome-efficiency rule in Section~4. As with episode-level data, we
remove malformed positive samples from local pairs.

\paragraph{Sampling budget.}
The rollout hyperparameters above are chosen based on preliminary runs.
We set episode-level collection to $K=10$, and local collection to two groups of top-3 high-entropy positions with \(K=4\) rollout at each branch point, so that episode-level and local collection produce comparable numbers and proportions of valid preference pairs under a similar rollout-time budget.
This avoids having one type of preference signal dominate the training data and allows the global outcome signal and the local invoke/skip signal to be combined more evenly.

\subsection{Training Rounds and Loss Masking}

We iteratively collect preference data with the previous round's model and train the next model only on that round's pairs, without accumulating data across rounds. ALFWorld-8B runs three SelSkill rounds from \textbf{RL-Init}, while ALFWorld-4B and BFCL run two rounds.

We stop at these points as valid preference pairs become sparse. One additional round reduces SR from 86.7 to 85.2, 77.3 to 73.4, and 24.2 to 22.6, respectively, consistent with preference-data saturation and overfitting.

For local pairs, we mask environment observations, skill returns, tool outputs, and tokens outside the first $n=3$ post-branch assistant turns. All training uses full-parameter fine-tuning with the hyperparameters in Table~\ref{tab:app_experimental_settings}.

\subsection{Compute Cost}

All runs use 8 NVIDIA A100 80GB GPUs. Table~\ref{tab:compute_cost} reports wall-clock costs; SelSkill's additional cost is offline and requires no counterfactual branching at inference time.

\begin{table}[t]
\centering
\footnotesize
\setlength{\tabcolsep}{3.2pt}
\renewcommand{\arraystretch}{1.05}
\begin{tabular}{@{}lrrrr@{}}
\toprule[1.5pt]
\textbf{Setting} & \textbf{GRPO} & \textbf{Rollout} & \textbf{Train} & \textbf{Total} \\
\midrule
ALFWorld 8B & 30.0 & 11.7 & 2.7 & \textbf{44.4} \\
ALFWorld 4B & 27.0 & 7.8 & 1.2 & \textbf{36.0} \\
BFCL 14B    & ---  & 4.6 & 1.8 & \textbf{6.4} \\
\midrule
\multicolumn{5}{@{}l@{}}{GRPO comparison (without / with skill): 27.0 / 64.0 h} \\
\bottomrule[1.5pt]
\end{tabular}
\caption{\textbf{Wall-clock time (hours) for the main experiments.}}
\label{tab:compute_cost}
\vspace{-8pt}
\end{table}

\section{Trajectory Analysis}
\label{app:trajectory_cases}

We illustrate three invocation patterns: context pollution from mistimed calls, failures from unmet preconditions, and successful calls made at an appropriate state.

\paragraph{BFCL context pollution.}
\label{app:case_bfcl}

Table~\ref{tab:case_bfcl} shows that invoking \texttt{place\_stock\_order} introduces a live price that overrides the user's specified limit price and derails an otherwise correct BFCL trajectory.

\begin{table*}[t!]
\small
\centering
\setlength{\tabcolsep}{4pt}
\renewcommand{\arraystretch}{1.08}
\begin{tabular}{@{}c p{0.445\linewidth}p{0.445\linewidth}@{}}
\toprule[1.5pt]
\textbf{Turn}
& \textbf{Without skill} \textcolor{teal}{(\checkmark)}
& \textbf{With skill} \textcolor{red}{(\texttimes)} \\
\midrule
1
& \texttt{place\_order(price=150, amount=100)} \newline
  $\to$ order 12446 \textcolor{teal}{\checkmark}
& \textbf{Skill(\texttt{place\_stock\_order})} \newline
  $\to$ \texttt{get\_stock\_info(AAPL)} $\to$ \$227.16 \newline
  $\to$ \texttt{place\_order(price=227.16, amount=100)} \newline
  $\to$ insufficient balance \textcolor{red}{\texttimes} \\
\midrule
2
& \texttt{get\_order\_details(12446)} \newline
  $\to$ correct recent order \textcolor{teal}{\checkmark}
& \texttt{get\_order\_history()} \newline
  $\to$ old completed order 12345 \newline
  \texttt{get\_order\_details(12345)} \textcolor{red}{\texttimes} \\
\midrule
3
& \texttt{cancel\_order(12446)} \textcolor{teal}{\checkmark}
& \texttt{cancel\_order(12345)} \newline
  $\to$ already completed \textcolor{red}{\texttimes} \\
\midrule
4
& \texttt{trading\_logout()} \textcolor{teal}{\checkmark}
& \texttt{trading\_logout()} \textcolor{teal}{\checkmark} \\
\bottomrule[1.5pt]
\end{tabular}
\caption{\textbf{Context pollution in BFCL.}
A mistimed skill call introduces an additional live price into the context, causing the model to override the user-specified limit price and derail subsequent execution.}
\label{tab:case_bfcl}
\end{table*}

\paragraph{ALFWorld premature call.}
\label{app:case_alfworld_false}

Table~\ref{tab:case_alfworld_false} shows a premature \texttt{examine\_with\_light} call before the agent acquires the desk lamp, causing repeated failures until the step budget is exhausted.

\begin{table*}[t!]
\small
\centering
\setlength{\tabcolsep}{5pt}
\renewcommand{\arraystretch}{1.08}
\begin{tabular}{@{}p{0.46\linewidth}p{0.46\linewidth}@{}}
\toprule[1.5pt]
\textbf{Without skill} \textcolor{teal}{(\checkmark)}
& \textbf{With skill} \textcolor{red}{(\texttimes)} \\
\midrule
\texttt{go to sidetable 1} \newline
\texttt{examine alarmclock 1} \newline
\texttt{go to desk 1} \newline
\texttt{pick up desklamp 1} \newline
\texttt{go to sidetable 1} \newline
\texttt{use desklamp 1 with alarmclock 1} \newline
$\to$ task complete \textcolor{teal}{\checkmark}
&
\texttt{go to sidetable 1} \newline
\texttt{examine alarmclock 1} \newline
\textbf{Skill(\texttt{examine\_with\_light}, alarmclock 1)} \newline
$\to$ ``Nothing happens.'' \textcolor{red}{\texttimes} \newline
\textbf{Skill(\texttt{examine\_with\_light}, alarmclock 1)} \newline
$\to$ ``Nothing happens.'' \textcolor{red}{\texttimes} \newline
\texttt{go to desk 1} \newline
\textbf{Skill(\texttt{examine\_with\_light}, alarmclock 1)} \newline
$\to$ ``Nothing happens.'' \textcolor{red}{\texttimes} \newline
\emph{repeats failing skill attempts} \newline
$\to$ step budget exhausted \textcolor{red}{\texttimes} \\
\bottomrule[1.5pt]
\end{tabular}
\caption{\textbf{Premature invocation in ALFWorld.}
The model invokes \texttt{examine\_with\_light} before acquiring the lamp, so the skill precondition is not satisfied.}
\label{tab:case_alfworld_false}
\end{table*}

\paragraph{ALFWorld correct call.}
\label{app:case_alfworld_correct}

Table~\ref{tab:case_alfworld_correct} shows a successful \texttt{heat\_object} call after the egg has been acquired and the skill precondition is satisfied.

\begin{table*}[t!]
\small
\centering
\setlength{\tabcolsep}{5pt}
\renewcommand{\arraystretch}{1.08}
\begin{tabular}{@{}p{0.46\linewidth}p{0.46\linewidth}@{}}
\toprule[1.5pt]
\textbf{With skill}  \textcolor{teal}{(\checkmark)}
& \textbf{Without skill} \textcolor{red}{(\texttimes)} \\
\midrule
\texttt{go to countertop 1} \newline
\texttt{go to countertop 2} \newline
\texttt{go to cabinet 1} \newline
\texttt{go to fridge 1} \newline
\texttt{open fridge 1} \newline
\texttt{take egg 3 from fridge 1} \newline
\textbf{Skill(\texttt{heat\_object})} \newline
$\to$ go to microwave \newline
$\to$ put egg in microwave \newline
$\to$ heat egg \newline
$\to$ retrieve hot egg \newline
\texttt{go to fridge 1} \newline
\texttt{move egg 3 to fridge 1} \newline
$\to$ task complete \textcolor{teal}{\checkmark}
&
\texttt{go to countertop 1} \newline
\texttt{go to countertop 2} \newline
\texttt{go to fridge 1} \newline
\texttt{go to stoveburner 1} \newline
\texttt{go to fridge 1} \newline
\texttt{go to sinkbasin 1} \newline
\emph{repeats navigation among countertops, stove burners, microwave, cabinets, sink, coffeemachine, and garbage can} \newline
\texttt{go to microwave 1} \newline
\emph{visits microwave but never executes the heat sequence} \newline
$\to$ step budget exhausted \textcolor{red}{\texttimes} \\
\bottomrule[1.5pt]
\end{tabular}
\caption{\textbf{Correct invocation in ALFWorld.}
For a task requiring a heated egg, SelSkill invokes \texttt{heat\_object} only after the egg has been picked up, when the skill precondition is satisfied.}
\label{tab:case_alfworld_correct}
\end{table*}
\end{document}